\documentclass{article}
\pdfoutput=1

\usepackage{my_style}

\usepackage[utf8]{inputenc} % allow utf-8 input
\usepackage[T1]{fontenc}    % use 8-bit T1 fonts
\usepackage{times}

\usepackage[round]{natbib}
\usepackage{graphicx}
\usepackage{rotating}
\usepackage{array}
\usepackage{algorithm}
\usepackage{algorithmic}
\usepackage[labelfont=bf]{caption}
\usepackage{subcaption}
\usepackage{amsmath,amsfonts,amssymb,bbm,epsfig,bm}

\usepackage{theorem}
\newcommand{\BlackBox}{\rule{1.5ex}{1.5ex}}  % end of proof

\newcommand{\cbr}[1]{\left\{#1\right\}}

\renewcommand{\algorithmicrequire}{\textbf{Input:}}
\renewcommand{\algorithmicensure}{\textbf{Output:}}

\newcolumntype{L}[1]{>{\raggedright\arraybackslash}p{#1}}
\newcolumntype{R}[1]{>{\raggedleft\arraybackslash}p{#1}}
\newcolumntype{C}[1]{>{\centering\let\newline\\\arraybackslash\hspace{0pt}}m{#1}}
\newcolumntype{?}{!{\vrule width 1pt}}
\makeatletter
\newcommand{\thickhline}{%
    \noalign {\ifnum 0=`}\fi \hrule height 1pt
    \futurelet \reserved@a \@xhline
}
\newcolumntype{"}{@{\hskip\tabcolsep\vrule width 1pt\hskip\tabcolsep}}
\makeatother
\newcommand{\intset}[1]{\cbr{1..n}}

\usepackage[colorlinks,pdfpagelabels,plainpages=false]{hyperref}
\usepackage{rotating}

\usepackage{xcolor}
\definecolor{dark-red}{rgb}{0.4,0.15,0.15}
\definecolor{dark-blue}{rgb}{0.15,0.15,0.4}
\definecolor{medium-blue}{rgb}{0,0,0.5}
\hypersetup{
   colorlinks, linkcolor={dark-blue},
   citecolor={dark-blue}, urlcolor={medium-blue}
}

\usepackage{booktabs}
\usepackage{bm}
\usepackage{amsmath}
\usepackage{amssymb}
\usepackage{amsfonts}
\usepackage{mathtools}
\usepackage{comment}

\newcommand{\mbf}[1]{{\boldsymbol{\mathbf{#1}}}}
\renewcommand{\bm}{\mbf}

\usepackage{color}

\usepackage{parskip}

\usepackage{multirow}
\usepackage{titlefoot}
\usepackage{gensymb}

\title{\bf Texar: A Modularized, Versatile, and Extensible Toolkit for Text Generation}

\author{
\normalsize
Zhiting Hu$^*$,~~
Haoran Shi,~~
Bowen Tan,~~
Wentao Wang,~~
Zichao Yang,\\
\normalsize
Tiancheng Zhao,~
Junxian He,~
Lianhui Qin,~
Di Wang,~
Xuezhe Ma,~
Zhengzhong Liu,
\\
\normalsize
Xiaodan Liang,~
Wangrong Zhu,~
Devendra Singh Sachan,~
Eric P. Xing\\
\normalsize
Carnegie Mellon University,~~ Petuum Inc.\\
\normalsize
{\tt zhitinghu@gmail.com}$^*$
}

\begin{document}
\date{}

\maketitle

\begin{abstract}

\begin{sloppypar}
We introduce Texar, an open-source toolkit aiming to support the broad set of \emph{text generation} tasks that transform any inputs into natural language, such as machine translation, summarization, dialog, content manipulation, and so forth. With the design goals of modularity, versatility, and extensibility in mind, Texar extracts common patterns underlying the diverse tasks and methodologies, creates a library of highly reusable modules, and allows arbitrary model architectures and algorithmic paradigms. In Texar, model architecture, inference, and learning processes are properly decomposed. Modules at a high concept level can be freely assembled and plugged in/swapped out. The toolkit also supports a rich set of large-scale pretrained models.
Texar is thus particularly suitable for researchers and practitioners to do fast prototyping and experimentation. The versatile toolkit also fosters technique sharing across different text generation tasks. Texar supports both TensorFlow and PyTorch, and is released under Apache~License~2.0 at \url{https://www.texar.io}.
\end{sloppypar}

\end{abstract}

\section{Introduction}\label{sec:intro}

Text generation spans a broad set of natural language processing tasks that aim to generate natural language from input data or machine representations. Such tasks include machine translation~\citep{brown1990statistical,vaswani2017attention}, dialog systems~\citep{williams2007partially,serban2016building,tang2019target}, text summarization~\citep{hovy1998automated,see2017get}, data description~\citep{wiseman2017challenges,li2018hybrid}, text paraphrasing and manipulation~\citep{hu2017controllable,madnani2010generating,lin2019toward}, image captioning~\citep{vinyals2015show,karpathy2015deep}, and more. Recent years have seen rapid progress of this active area, in part due to the integration of modern deep learning approaches in many of the tasks. On the other hand, considerable research efforts are still needed in order to improve techniques and enable real-world applications.

\begin{figure*}%[t]
  \centering 
  \includegraphics[width=0.9\linewidth]{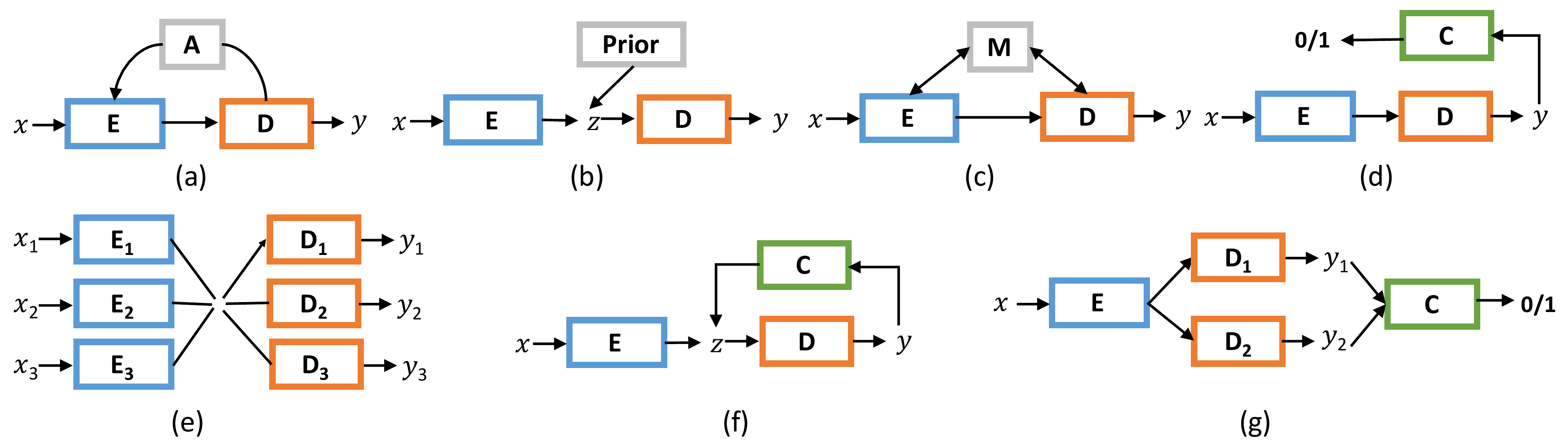}
  \vspace{-10pt}
  \caption{An example of various model architectures in recent text generation literatures. \texttt{E} denotes encoder, \texttt{D} denotes decoder, \texttt{C} denotes classifier (i.e., binary discriminator). {\bf (a): } The canonical encoder-decoder, sometimes with attention \texttt{A}~\citep{sutskever2014sequence,bahdanau2014neural,luong2015effective,vaswani2017attention} or copy mechanisms~\citep{gu2016incorporating,vinyals2015pointer,gulcehre2016pointing}. {\bf (b): } Variational encoder-decoder~\citep{bowman2015generating,yang2017improved}. {\bf (c): } Encoder-decoder augmented with external memory~\citep{sukhbaatar2015end,bordes2016learning}. {\bf (d): } Adversarial model using a binary discriminator \texttt{C}, with or without reinforcement learning~\citep{liang2017recurrent,zhang2017adversarial,yu2017seqgan}. {\bf (e): } Multi-task learning with multiple encoders and/or decoders~\citep{luong2015multi,firat2016multi}. {\bf (f): } Augmenting with cyclic loss~\citep{hu2017controllable}. {\bf (g): } Learning to align with adversary, either on samples $y$ or hidden states~\citep{lamb2016professor,lample2017unsupervised,shen2017style}.} 
  \label{fig:archs}
% \end{figure*}
% \begin{figure*}[t]
%   \centering 
  \includegraphics[width=0.75\linewidth]{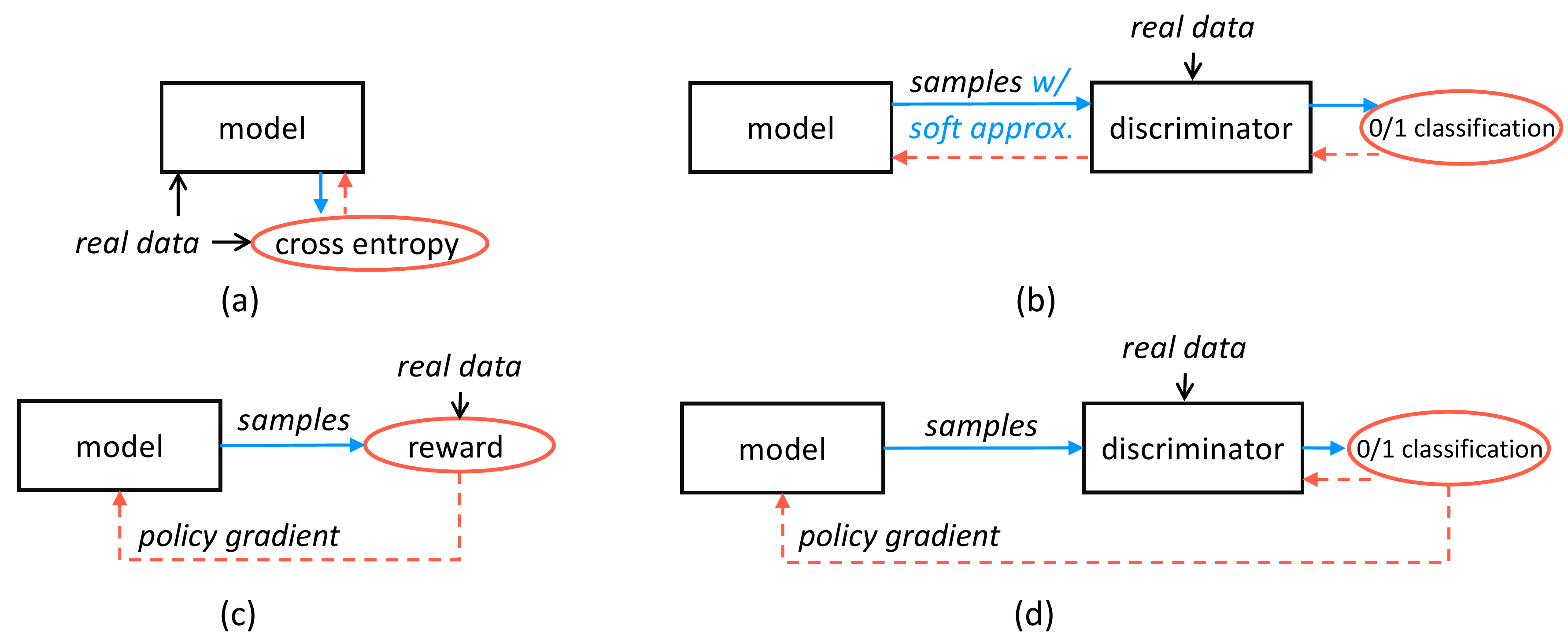}
  \vspace{-10pt}
  \caption{An example of various learning paradigms for text generation models. Inference processes are denoted as blue solid arrows; learning signals and gradient propagation are denoted as red dashed arrows. {\bf (a): } Maximum likelihood learning with a cross-entropy loss~\citep{mikolov2010recurrent}. {\bf (b): } Adversarial learning that propagates gradient through samples with continuous approximation ({\it soft approx.})~\citep{hu2017controllable,yang2018unsupervised}. A discriminator is trained jointly with the target model. {\bf (c): } Reinforcement learning with a specified reward function (e.g., BLEU score) and policy gradient~\citep{ranzato2015sequence,rennie2017self}. {\bf (d): } Combination of adversarial learning and reinforcement learning by using a learnable discriminator as the reward function and updating the target model with policy gradient~\citep{fedus2018maskgan,yu2017seqgan}. Other learning algorithms can include reward-augmented maximum likelihood~\citep{norouzi2016reward}, learning-to-search~\citep{hal2009search,wiseman2016sequence}, interpolation algorithm~\citep{tan2018connecting}, etc.
  %In practice, the different learning paradigms are usually used together. For example, a model can first be pre-trained with maximum likelihood learning, and then be fine-tuned with either adversarial or reinforcement learning. 
  } 
  \label{fig:archs-alg}
\end{figure*}

A few remarkable open-source toolkits have been developed in support of text generation applications (sec~\ref{sec:related}). Those toolkits, however, are largely designed for one or a small number of specific tasks, particularly machine translation~\citep[e.g.,][]{britz2017massive,klein2017opennmt} and conversation systems~\citep[e.g.,][]{miller2017parlai}, and usually support a narrow set of machine learning algorithms such as supervised learning. Emerging new applications and approaches instead are often developed by individual teams in a more ad-hoc manner, which can easily result in hard-to-maintain custom code and duplicated efforts across the disjoint projects.

The variety of text generation tasks indeed have many common properties. For instance, two central goals are shared across most of them, namely 1) generating well-formed, grammatical, and readable text, and 2) realizing in the generated text all desired information inferred from the inputs. To this end, a set of key techniques are increasingly widely-used, such as neural encoder-decoders~\citep{sutskever2014sequence}, attention~\citep{bahdanau2014neural,luong2015effective,vaswani2017attention}, memory networks~\citep{sukhbaatar2015end}, adversarial methods~\citep{goodfellow2014generative,lamb2016professor}, reinforcement learning~\citep{ranzato2015sequence,bahdanau2016actor,tan2018connecting}, structured supervision~\citep{hu2018deep,yang2018unsupervised}, as well as optimization techniques, data pre-processing and result post-processing procedures, evaluations, etc. These techniques are often combined together in various ways to tackle different problems. Figures~\ref{fig:archs} and~\ref{fig:archs-alg} summarize examples of various model architectures and learning paradigms, respectively.  

It is therefore highly desirable to have an open-source platform that unifies the development of the diverse yet closely-related applications, backed with clean and consistent implementations of the core algorithms. Such a platform would enable reuse of common components and functionalities; standardize design, implementation, and experimentation; foster reproducibility; and, importantly, encourage technique sharing among different text generation tasks so that an algorithmic advance developed for a specific task can quickly be evaluated and generalized to many other tasks.
%other tasks can immediately share the advances made in any specific task.

\begin{figure*}[t]
  \centering 
  \includegraphics[width=0.95\linewidth]{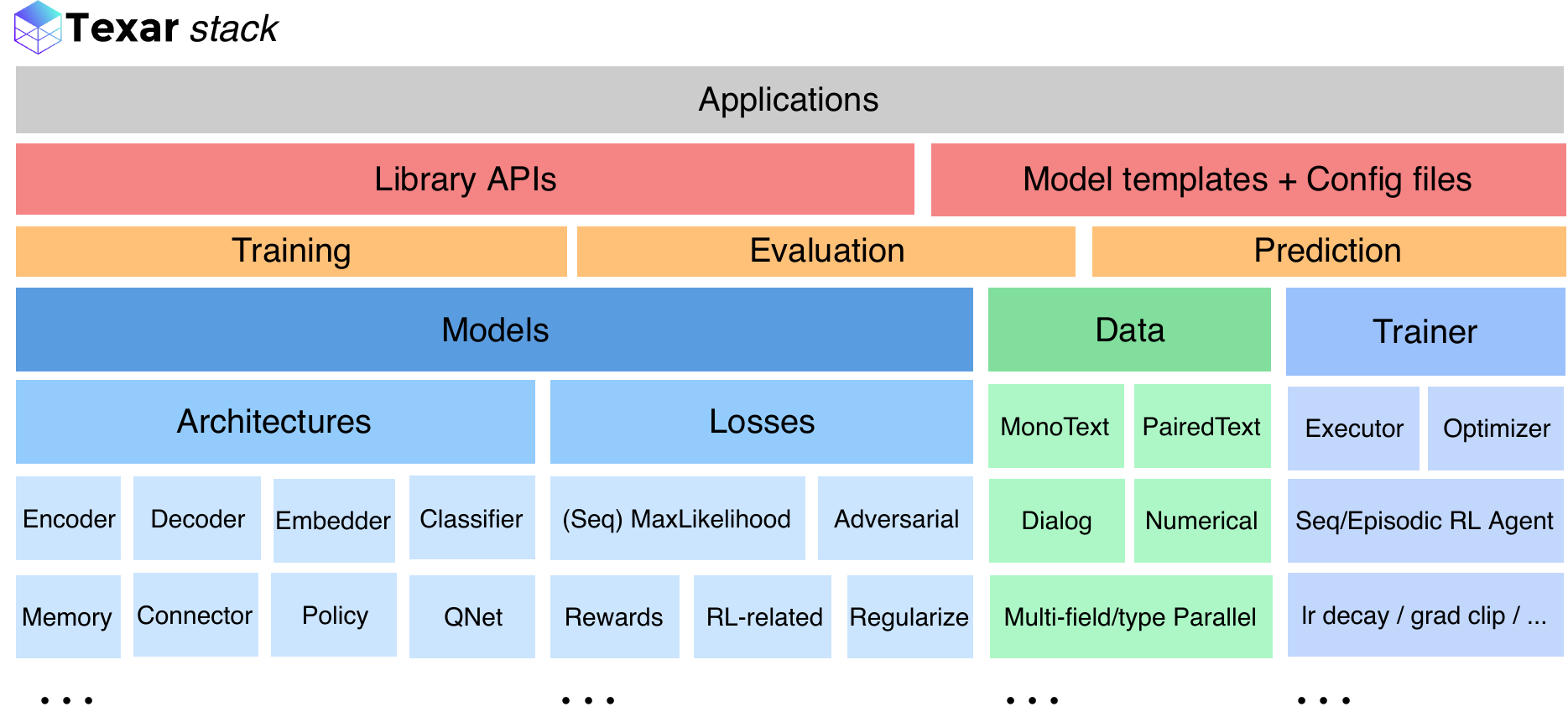}
  \vspace{-8pt}
  \caption{The stack of main modules and functionalities in Texar. Many of the low-level components are omitted.} 
  \label{fig:stack}
  \vspace{-8pt}
\end{figure*}

We introduce {\it Texar}, a general-purpose text generation toolkit aiming to support popular and emerging applications in the field, by providing researchers and practitioners a unified and flexible framework for building their models. 
Texar has two versions, building upon TensorFlow (\url{tensorflow.org}) and PyTorch (\url{pytorch.org}), respectively, with the same uniform design.

Underlying the core of Texar's design is principled anatomy of extensive machine learning models and algorithms~\citep{hu2017unifying,tan2018connecting}, which subsumes the diverse cases in Figures~\ref{fig:archs},~\ref{fig:archs-alg} and beyond, enabling a unified formulation and consistent implementation. Texar emphasizes three key properties:

\begin{itemize}
\item {\bf Versatility}: Texar contains a wide range of features and functionalities for 1) arbitrary model architectures as a combination of encoders, decoders, embedders, discriminators, memories, and many other modules; and 2) different modeling and learning paradigms such as sequence-to-sequence, probabilistic models, adversarial methods, and reinforcement learning. Based on these, both workhorse and cutting-edge solutions to the broad spectrum of text generation tasks are either already included or can be easily constructed.
\item {\bf Modularity}:
Texar is designed to be highly modularized, by decoupling solutions of diverse tasks into a set of highly reusable modules. Users can construct their model at a high conceptual level just like assembling building blocks. It is convenient to plug in or swap out modules, configure rich options of each module, or even switch between distinct modeling paradigms. For example, switching between maximum likelihood learning and reinforcement learning involves only minimal code changes (e.g., Figure~\ref{fig:switch}). Modularity makes Texar particularly suitable for fast prototyping and experimentation.
\item {\bf Extensibility}:
The toolkit provides interfaces of multiple functionality levels, ranging from simple configuration files to full library APIs. Users of different needs and expertise are free to choose different interfaces for appropriate programmability and internal accessibility. The library APIs are fully compatible with the native TensorFlow/PyTorch interfaces, which allows seamless integration of user-customized modules, and enables the toolkit to take advantage of the vibrant open-source community by effortlessly importing any external components as needed.
\end{itemize}

Furthermore, Texar puts much emphasis on well-structured high-quality code of uniform design patterns and consistent styles, along with clean documentations and rich tutorial examples. Other useful features such as distributed GPU training are also supported.

%In the following, we first give a brief overview of related open-source tools (section~\ref{sec:related}). We then provide details of the toolkit structure and design (section~\ref{sec:texar_design}), highlighting the principled unified anatomy of diverse learning algorithms and models that underlies the modularized design of the toolkit and enables free assembly of modules as well as effortless switch between learning paradigms. We further conduct case studies to demonstrate that Texar reduces development efforts and helps study of technique sharing across different text generation tasks (section~\ref{sec:exp}).
%
%To demonstrate the use of the toolkit and its advantages, we perform extensive experiments and cases studies (section~\ref{sec:exp}), including generalizing the state-of-the-art machine translation model to multiple text generation tasks, investigating different algorithms for language modeling, and implementing composite neural architectures beyond conventional encoder-decoder for text style transfer. 
%All are easily realized with the versatile toolkit.

%Texar is released at: [{\it URL omitted for blind review}]. We attach the anonymized code of the toolkit in supplementary materials.
%For the sake of blind review, we hold an \emph{anonymous} GitHub repository containing the code of the toolkit: \url{https://anonymous.4open.science/repository/6fbe3487-be4c-4d09-83a0-23cc16914e48}

\begin{figure}[t]
  \centering 
  \includegraphics[width=0.6\linewidth]{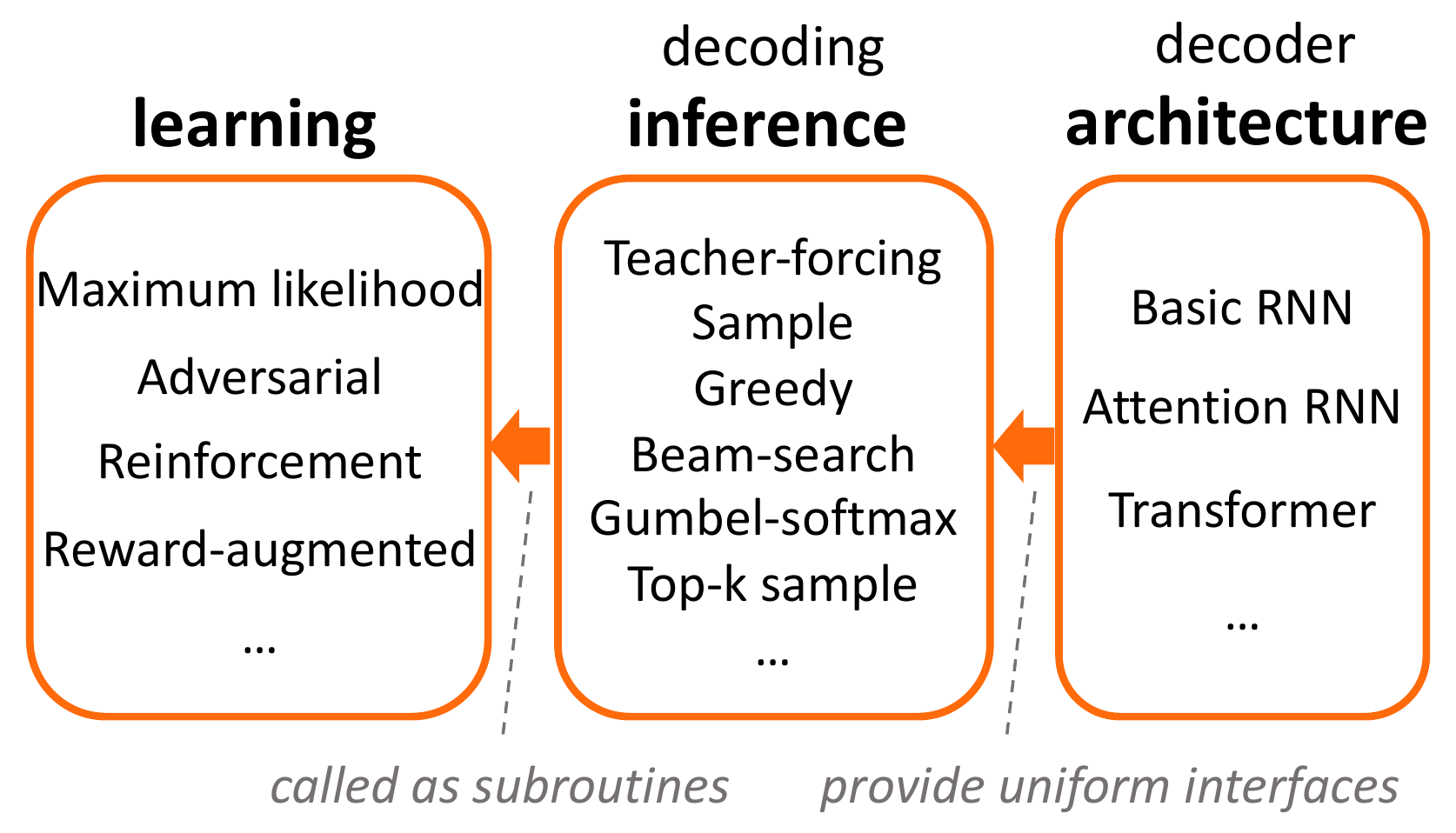}
  \vspace{-10pt}
  \caption{The learning--inference--architecture anatomy in Texar, taking decoder for example. Specifically, a sequence decoder in a model can have an arbitrary architecture (e.g., basic RNN decoder); all architectures expose uniform interfaces for specifying one of the tens of decoding strategies (e.g., teacher-forcing decoding) to generate samples or infer probabilities; a learning procedure repeated calls arbitrary specified inference procedure during training. Learning can be totally agnostic to the model architecture.} 
  \label{fig:anatomy}
  %\vspace{-6pt}
\end{figure}

\section{Structure and Design}\label{sec:texar_design}

Figure~\ref{fig:stack} shows the stack of main modules and functionalities in Texar. Building upon the lower level deep learning platforms, Texar provides a comprehensive set of building blocks for model construction, training, evaluation, and prediction. %Texar is designed with the goals of \emph{versatility, modularity, and extensibility} in mind.
In the following, we first present the design principles that lead to the attainment of these goals (sec~\ref{sec:design}), and then describe the detailed structure of Texar with running examples to demonstrate the properties of the toolkit (sec~\ref{sec:assemble}-\ref{sec:interface}).

\subsection{The Design of Texar}\label{sec:design}
The broad variation of the many text generation tasks and the fast-growing new models and algorithms have posed unique challenges to designing a versatile toolkit. We tackle the challenges by principally decomposing the modeling and experimentation pipeline, developing an extensive set of ready-to-assemble modules, and providing user interfaces of varying abstract levels.

{\bf Unified Anatomy. }
We break down the complexity of the rich text generation tasks into three dimensions of variations, namely, the varying data types and formats for different use cases, the arbitrary combinational model architectures and associated inference procedures (e.g., Figure~\ref{fig:archs}), and the distinct learning algorithms such as maximum likelihood learning, reinforcement learning, and combinations thereof (e.g., Figure~\ref{fig:archs-alg}). 
Within the unified abstraction, all learning paradigms are each specifying one or multiple loss functions (e.g., cross-entropy loss, policy gradient loss), along with an optimization procedure that improves the losses: 
%In particular, prior literatures and systems have usually viewed the various algorithms as distinct learning paradigms and having different formulations. For example, maximum likelihood learning aims to maximize the data log-likelihood, while reinforcement learning is devoted to interacting with the environment and improving from the feedback. Here, in contrast, we adopt a unified algorithmic abstraction in which all the paradigms are each specifying one or multiple loss functions (e.g., cross-entropy loss, policy gradient loss), along with an optimization procedure that improves the losses: 
%We begin with a high-level decomposition of model construction and learning pipeline. A deep neural model is typically learned with the following abstract procedure:
\begin{equation}
\begin{split}
\min\nolimits_\theta \mathcal{L}(f_\theta, D)
\end{split}
\end{equation}
where $\bm{f}_\theta$ is the model that defines the model architecture and the inference procedure; $\bm{D}$ is the data; $\bm{\mathcal{L}}$ is the learning objectives (losses); and $\bm{\min}$ denotes the optimization procedure. Note that the above can have multiple losses imposed on different model parts (e.g., adversarial learning). 

\begin{figure*}[t]
  \centering 
  \includegraphics[width=0.95\linewidth]{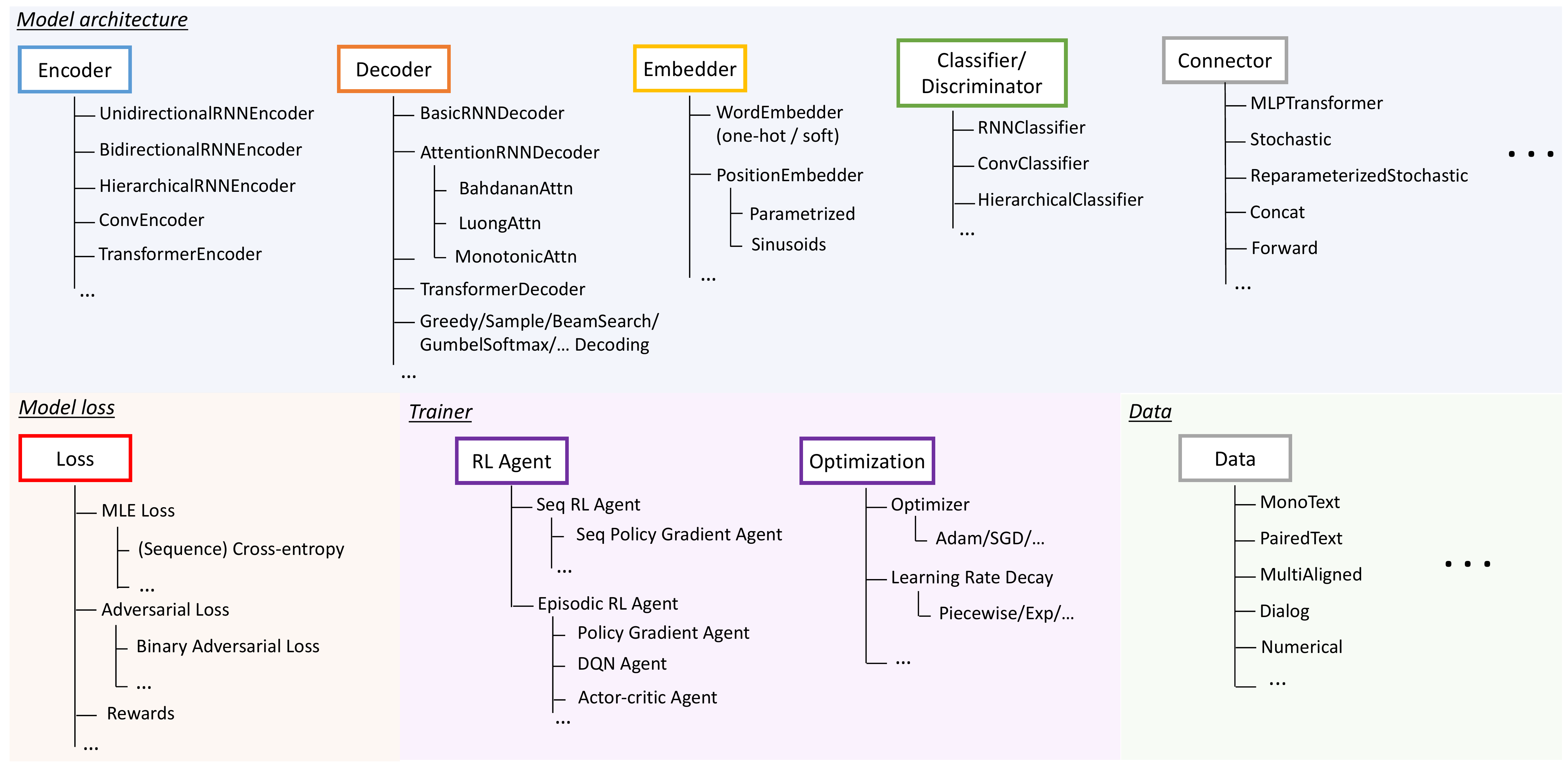}
  \vspace{-6pt}
  \caption{The catalog of a subset of modules for model construction and learning. Other modules, such as memory network modules, and those for evaluation and prediction, are omitted due to space limitations. 
} 
  \label{fig:catalog}
  %\vspace{-10pt}
\end{figure*}

Further, as illustrated in Figure~\ref{fig:anatomy}, we decouple learning, inference, and model architecture to the maximum extent, forming abstraction layers of {\bf learning -- inference -- architecture}, as illustrated in Figure~\ref{fig:anatomy}. That is, different architectures implement the same set of inference procedures and provide the same interfaces, so that learning algorithms can call proper inference procedures as subroutines while staying agnostic to the underlying architecture and implementation details. For example, maximum likelihood learning uses teacher-forcing decoding~\citep{mikolov2010recurrent}; a policy gradient algorithm can invoke stochastic or greedy decoding~\citep{ranzato2015sequence}; and adversarial learning can use either stochastic decoding for policy gradient-based updates~\citep{yu2017seqgan} or Gumbel-softmax reparameterized decoding~\citep{jang2016categorical} for direct gradient back-propagation. Users can effortlessly switch between different learning algorithms for the same model, by simply specifying the corresponding inference strategy and plugging into the new learning module, without adapting the model architecture (see section~\ref{sec:plugin} for a running example). 

The unified anatomy has underlay the strong modularity of the toolkit. It helps maximize the opportunities for reuse, enable free combinations of different parts, and greatly improve the cleanness of code structure.

%Texar is designed to properly decouple the four elements, and allow free combinations of them through uniform interfaces. Such design has underlay the strong modularity of the toolkit.

\begin{figure*}[t]
  \centering 
  \includegraphics[width=\linewidth]{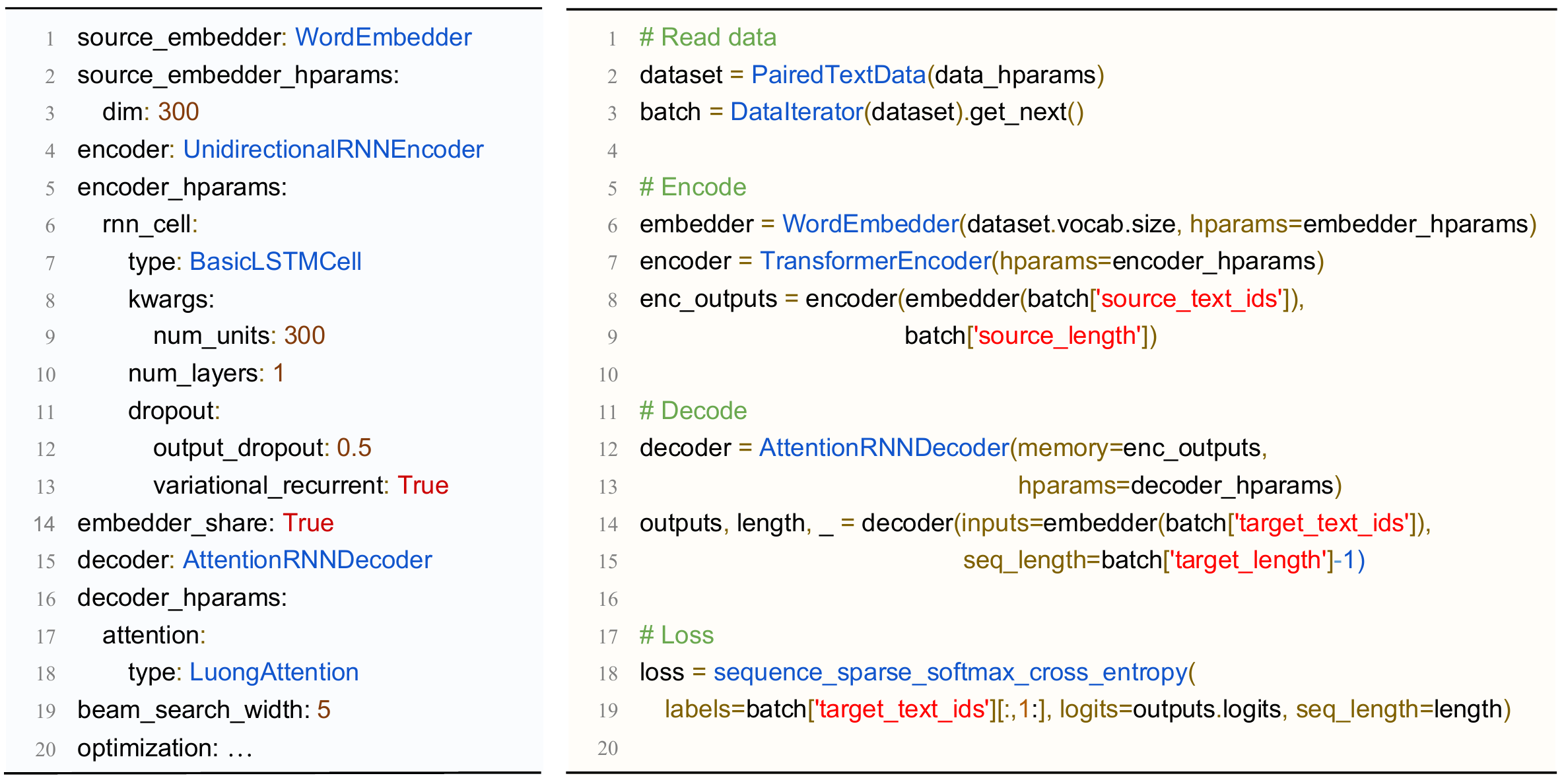}
  \vspace{-15pt}
  \caption{Two ways of specifying an attentional sequence-to-sequence model. {\bf Left: }Snippet of an example YAML configuration file of the sequence-to-sequence model template. Only those hyperparameters that the user concerns are specified explicitly in the particular file, while the remaining many hyperparameters can be omitted and will take default values. {\bf Right: }Python code assembling the sequence-to-sequence model using the Texar library APIs. Modules are created as Python objects, and then called as functions to add TensorFlow ops to the computation graph and return output tensors. Other code such as optimization is omitted. } 
  \label{fig:attn_code}
  %\vspace{-10pt}
\end{figure*}

% In particular, the decomposition of model architecture and inference (i.e., $f_\theta$) from losses and learning has greatly improved the cleanness of the code structure and the opportunities for reuse. For example, a sequence decoder can focus solely on performing different decoding (inference) schemes, such as decoding with ground truths (teacher-forcing), and greedy, stochastic, and beam-search decoding, etc. Different learning algorithms then call different schemes as a subroutine in the learning procedure---for example, maximum likelihood learning uses decoding with ground truths~\citep{mikolov2010recurrent}, a policy gradient algorithm can use stochastic decoding~\citep{ranzato2015sequence}, and an adversarial learning can use either stochastic decoding for policy gradient-based updates~\citep{yu2017seqgan} or Gumbel-softmax reparameterized decoding~\citep{jang2016categorical} for direct gradient back-propagation (e.g., Figure~\ref{fig:switch}). 
%With unified abstractions, the decoder and the learning algorithms are agnostic to the implementation details of each other. 
%This also enables convenient switch between different learning algorithms for the same model, by simply changing the inference scheme and connecting to the new learning module, without adapting the model architecture (see sec~\ref{sec:plugin} for the example). 

{\bf Module Assembly. } 
The fast evolution of modeling and learning methodologies has led to sophisticated models that go beyond the canonical (attentional) sequence-to-sequence alike forms and introduce many new composite architectures. 
(Figure~\ref{fig:archs} summarizes several model architectures developed in recent literature for different tasks.) To versatilely support all these diverse approaches, we break down the complex models and extract a set of frequently-used modules (e.g., \texttt{encoders}, \texttt{decoders}, \texttt{embedders}, \texttt{classifiers}, etc). Figure~\ref{fig:catalog} shows the catelog of a subset of modules. Crucially, Texar allows free concatenation between these modules in order to assemble arbitrary model architectures. Such concatenation can be done by directly interfacing two modules, or through an intermediate \texttt{connector} module that provides general functionalities of shape transformation, reparameterization~\citep[e.g.,][]{kingma2013auto,jang2016categorical}, sampling, and others.

{\bf User Interfaces. }
It is crucial for the toolkit to be flexible enough to allow construction of simple and advanced models, while at the same time providing proper abstractions to relieve users from overly concerning about low-level implementations. 
In particular, Texar provides two major types of user interfaces: 1) YAML configuration files that instantiate pre-defined model templates, and 2) full Python library APIs. The former is simple, clean, straightforwardly understandable for non-expert users, and is also adopted by other toolkits~\citep{britz2017massive,klein2017opennmt}, while the latter allows maximal flexibility, full access to internal states, and essentially unlimited customizability. The libray APIs also provide interfaces at different abstract levels for key functionalities, allowing users to select and trade off between readily usability and customizability.
Examples are provided in the following sections. 

\subsection{Assemble Arbitrary Model Architectures}\label{sec:assemble}
Figure~\ref{fig:attn_code} shows an example of specifying an attentional sequence-to-sequence model through either a YAML configuration file (left panel), or concise Python code (right panel), respectively. 

\begin{figure*}[t]
  \centering 
  \includegraphics[width=\linewidth]{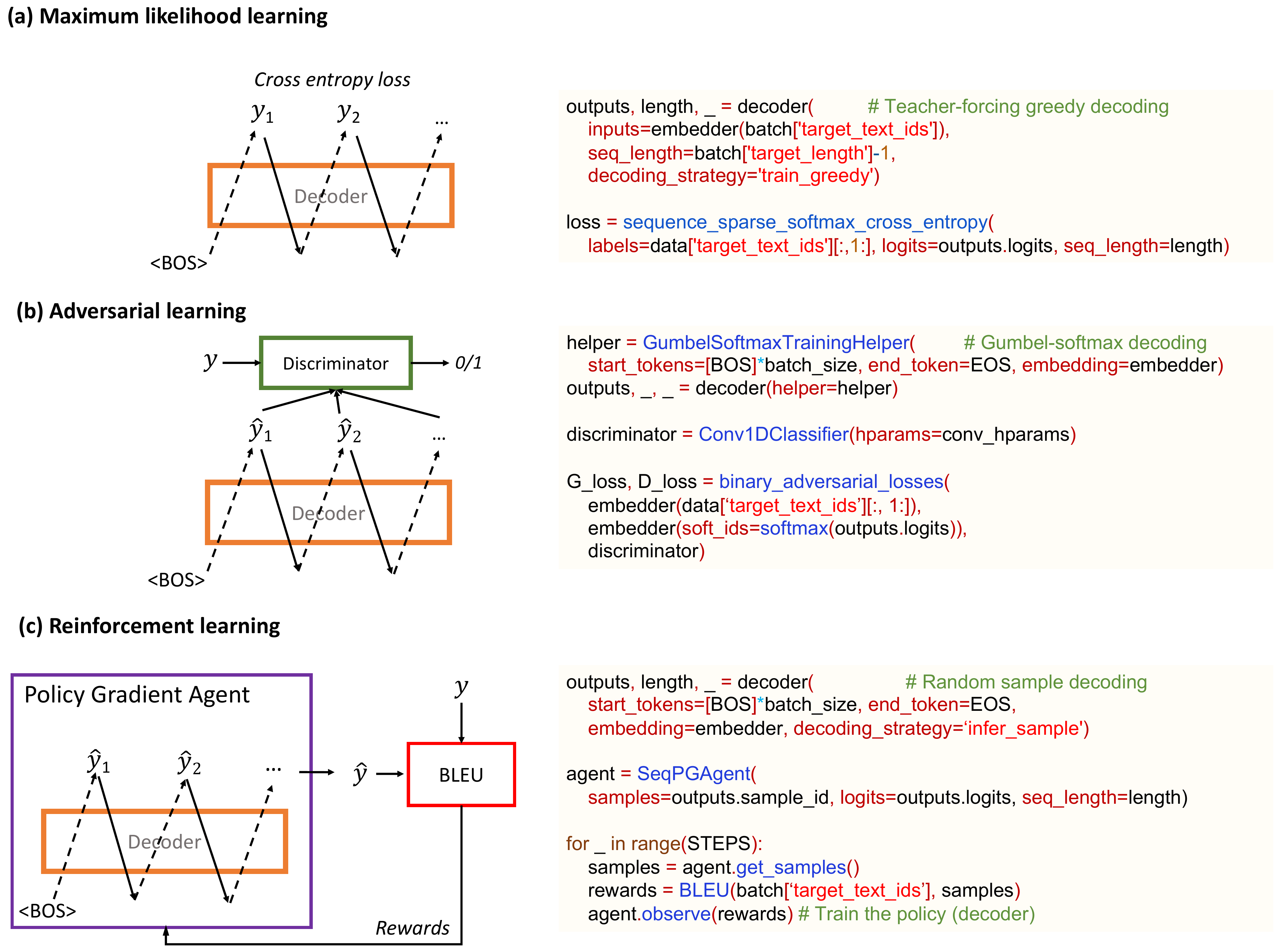}
  \vspace{-18pt}
  \caption{Switching between different learning paradigms of a decoder involves only modification of Line.14-19 in the right panel of Figure~\ref{fig:attn_code}. In particular, the same decoder is called with different decoding strategies, and discriminator or reinforcement learning agent is added as needed with simple API calls.
  {\bf (Left): }Module structure of each paradigm; {\bf (Right): }The respective code snippets. For adversarial learning in (b), continuous Gumbel-softmax approximation~\citep{jang2016categorical} to the generated samples (with \texttt{GumbelSoftmaxTrainingHelper}) is used to enable gradient propagation from the discriminator to the decoder.} 
  \label{fig:switch}
  \vspace{-6pt}
\end{figure*}

\begin{itemize}
\item The configuration file passes hyperparameters to the model template which instantiates the model for subsequent training and evaluation (which are also configured through YAML). Text highlighted in blue in the figure specifies the names of modules to use. Module hyperparameters are specified under \texttt{*\_hparams} in the configuration hierarchy (for example, \texttt{source\_embedder\_hparams} for the \texttt{source\_embedder} module). Note that most of the hyperparameters have sensible default values, and users only have to specify a small subset of them. Hyperparameters taking default values can be omitted in the configuration file.
\item The library APIs enable users to efficiently build any desired pipelines at a high conceptual level, without worrying too much about the low-level implementations. Power users are also given the option to access the full internal states for native programming and low-level manipulations. 

Texar modules have multiple features for ease of use, including 1) {\it Convenient variable re-use}:  Each module instance creates its own sets of variables, and automatically re-uses its variables on subsequent calls. Hence TensorFlow {\it variable scope} is transparent to users; 2) {\it Configurable through hyperparameters}: Each module defines allowed hyperparameters and default values. Hyperparameter values are configured by passing the \texttt{hparams} argument to the module constructor, which precisely corresponds to the \texttt{*\_hparams} sections in YAML configuration files; 3) {\it PyTorch-alike function calls}: As in Figure~\ref{fig:attn_code}, after a module is created as an object, it can be called as a function which performs the module logic on input tensors and returns output tensors. Both the Texar TensorFlow and PyTorch versions have the same interfaces.
\end{itemize}

\subsection{Plug-in and Swap-out Modules}\label{sec:plugin}
Texar builds a shared abstraction of the broad set of text generation tasks. 
It is convenient to switch between different application contexts, or change from one modeling paradigm to another, by simply plugging in/swapping out a single or few modules, or even merely changing a configuration parameter, while keeping all other parts of the modeling pipeline unchanged. 

For example, given the base code of the sequence-to-sequence model in Figure~\ref{fig:attn_code} (right panel), Figure~\ref{fig:switch} illustrates how Texar can easily support switching between different learning algorithms, by changing only the relevant code snippet (i.e., Line.14--19 in Figure~\ref{fig:attn_code} right panel). In particular, Figure~\ref{fig:switch} shows three major learning paradigms, including maximum-likelihood based supervised learning, adversarial learning, and reinforcement learning, each of which invokes different decoding (inference) methods of the decoder, namely teacher-forcing decoding, Gumbel-softmax decoding, and random-sample decoding, respectively. 
%We can see all these changes are done by simply adjusting the API calls.
%only local modification of few lines of code is enough to achieve such change,
%and discriminator or reinforcement learning agent is added when needed, with simple API calls.
%In particular, the same decoder is called with different decoding modes (e.g., \texttt{train\_greedy}, \texttt{Gumbel-softmax}, and \texttt{infer\_sample}), 

The convenient module switch can be useful for fast exploration of different algorithms for a specific task, or quick experimentation of an algorithm's generalization on different tasks.

\subsection{Customize with Extensible Interfaces}\label{sec:interface}
Texar emphasizes heavily on extensibility, and allows easy addition of customized or external modules through various interfaces, without editing the Texar codebase.

With the YAML configuration file, users can directly insert their own modules by providing the Python importing path to the module. For example, to use an externally implemented RNN cell in the sequence-to-sequence model encoder, one can simply change Lines.6-9 in the left panel of Figure~\ref{fig:attn_code} to the following:
\begin{figure}[!h]
\vspace{-6pt}
  \centering 
  \includegraphics[width=0.5\columnwidth]{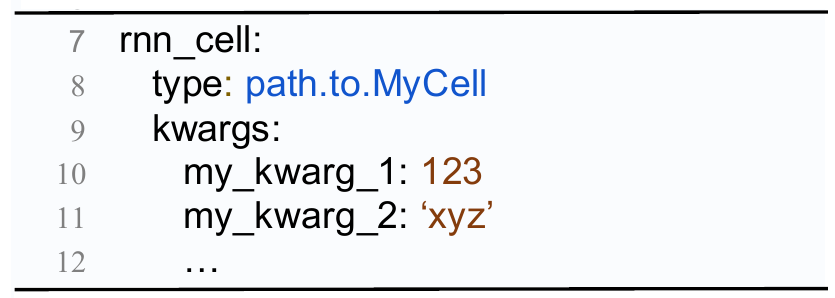}
  \vspace{-7pt}
\end{figure} 
\\
as long as the \texttt{MyCell} class is accessible by Python, and its interface is compatible to other parts of the model.

Incorporating customized modules with Texar library APIs is even more flexible and straightforward. As the library APIs are designed to be coherent with the native TensorFlow/PyTorch programming interfaces, any externally-defined modules can directly be combined with Texar components as needed.

\section{Case Studies}\label{sec:exp}
In this section, we conduct several case studies to show that Texar can greatly reduce implementation efforts and easily enable technique sharing among different tasks.
%In this section, we use Texar to conduct several case studies. We show that Texar enables easy replication of previous results, reduced implementation efforts compared to ad-hoc development, and \emph{technique sharing} that has been advantageously supported by the toolkit. 

% Specifically, thanks to the modularized design and versatile support of text generation applications, in section~\ref{sec:exp-one-tech}, we study the generalization of a {\it single technique on different tasks}. We are able to
% %We conduct case studies on \emph{technique sharing} that is advantageously supported by Texar: (1) We 
% deploy the state-of-the-art machine translation model on language modeling and conversation generation, and obtain improved performance over previous methods. In  section~\ref{sec:exp-one-task}, we instead study the use of {\it different techniques on a single task}, obtaining benchmark results of language modeling. Finally, in section~\ref{sec:exp-style}, we 
% %(2) We apply various model paradigms on the task of language modeling to compare the different methods. Besides, to further demonstrate the versatility of Texar, we
% show benchmark performance on the newly-emerging task of text style transfer, which involves composite neural architectures beyond conventional encoder-decoder.

\begin{table}[t]
\small
\centering
\begin{tabular}{l l l l}
\cmidrule[\heavyrulewidth]{1-4}
\multicolumn{4}{l}{{\bf Task:} VAE language modeling} \\ \cmidrule{1-4}
{\bf Dataset} & {\bf Metrics} & {\bf VAE-LSTM} & {\bf VAE-Transformer} \\ \cmidrule{1-4}
\multirow{2}{*}{Yahoo~\citep{yang2017improved}} & Test PPL & 68.31 & {\bf 61.26} \\
& Test NLL  & 337.36 & {\bf 328.67} \\ \cmidrule{1-4}
\multirow{2}{*}{PTB~\citep{bowman2015generating}} & Test PPL & 105.27 & {\bf 102.46} \\
& Test NLL  & 102.06 & {\bf 101.46} \\
\cmidrule[\heavyrulewidth]{1-4}
\end{tabular}
\vspace{-6pt}
\caption{Deployment of Transformer on the task of VAE language modeling~\citep{bowman2015generating}.
Both test set perplexity (PPL) and sentence-level negative log likelihood (NLL) are evaluated (The lower the better). The model with a Transformer decoder consistently outperforms the one with a conventional LSTM decoder. For fair comparison, both models have the same parameter size. %Notably, changing from the LSTM model to the Transformer model involves only modifying 3 lines of code.
%Both the LSTM and Transformer  decoders have around 6.3M free parameters.
}
\label{tab:transformer-lm}
%\vspace{-15pt}
\end{table}

\subsection{One Technique on Many Tasks: Transformer}\label{sec:exp-one-tech}

Transformer, as first introduced in~\citep{vaswani2017attention}, has greatly improved the machine translation results and created other successful models, such as BERT for text embedding~\citep{devlin2018bert}, GPT-2 for language modeling~\citep{radford2018language}, text infilling~\citep{zhu2019text}, etc. Texar supports easy construction of these models and fine-tuning pretrained weights. On Texar, we can easily deploy the Transformer components to various other tasks and get improved results.

%Transformer~\citep{vaswani2017attention} is a recently developed {\it machine translation} model that introduces a new {\it self-attention} technique and has achieved state-of-the-art performance.
%%Different from the widely-used attentional sequence-to-sequence models~\citep{bahdanau2014neural}, Transformer introduces a new \emph{self-attention} technique in which each generated token attends to all previously generated tokens. 
%It would be interesting to see how the cutting-edge technique can generalize to other text generation tasks beyond machine translation. With Texar, we deploy the Transformer decoder on two tasks, namely variational autoencoder (VAE) based language modeling~\citep{bowman2015generating} and conversation generation~\citep{serban2016building}. 

The first task we explore is the variational autoencoder (VAE) language modeling~\citep{bowman2015generating}. LSTM RNN has been widely-used in VAE for decoding sentences. We follow the experimental setting in previous work~\citep{bowman2015generating,yang2017improved}, and test two models, one with the traditional LSTM RNN decoder, and the other with the Transformer decoder. All other model configurations are the same in the two models. Notably, with Texar, changing the decoder from an LSTM to a Transformer is achieved by modifying only 3 lines of code.
Table~\ref{tab:transformer-lm} shows the results. We see that the Transformer VAE consistently improves over the conventional LSTM VAE, showing that the Transformer architecture can benefit tasks beyond machine translation. 

It is worth noting that, building the VAE language model (including data reading, model construction and optimization specification) on Texar uses only 70 lines of code (with the length of each line $<80$ chars). As a (rough) reference, a popular public TensorFlow code~\citep{vaelmtf} of the same model has used around 400 lines of code for the same part (without line length limit). 
%(Furthermore, Texar strictly limits the length of each line to $<80$ chars, while others typically do not.)

\begin{table}
\small
\centering
\begin{tabular}{l l l}
\cmidrule[\heavyrulewidth]{1-3}
\multicolumn{3}{l}{{\bf Task:} Conversation generation} \\\cmidrule{1-3}
{\bf Metrics} & {\bf HERD-GRU} & {\bf HERD-Transformer} \\
\cmidrule{1-3}
BLEU-3 prec  & 0.281 & \bf{0.289} \\
BLEU-3 recall & 0.256 & \bf{0.273} \\
\cmidrule{1-3}
BLEU-4 prec  & 0.228 & \bf{0.232} \\
BLEU-4 recall & 0.205 & \bf{0.214} \\
\cmidrule[\heavyrulewidth]{1-3}
\end{tabular}
\vspace{-6pt}
\caption{Comparison of Transformer decoder and GRU RNN decoder within the conversation model HERD~\citep{serban2016building} for response generation, on the Switchboard dataset~\citep{zhao2017learning}.}
\label{tab:transformer-dialog}
%\vspace{-15pt}
\end{table}

The second task is to generate a response given conversation history. Following~\citep{serban2016building}, the conversation history is encoded with a Texar module \texttt{HierarchicalRNNEncoder} which is followed with a decoder to generate the response. Similar as above, we study the performance of a Transformer decoder in comparison with a more conventional GRU RNN decoder.
%
%We use the popular hierarchical recurrent encoder-decoder model (HRED)~\citep{serban2016building} as the base model, which treats a conversation as a transduction task. The conversation history is seen as the source sequence and is modeled with a {\it hierarchical encoder}. Each utterance in the dialog history is first encoded with a word-level RNN. The resulting hidden states of the sequence of utterance are then encoded with an utterance-level RNN. We follow the experimental setting in~\citep{zhao2017learning}. In particular, the word-level RNN is set to be bidirectional and the utterance-level is unidirectional. Such configuration is easily implemented by setting the hyperparameters of the Texar module \texttt{HierarchicalRNNEncoder}. 
%Similar to the above task, we compare two models, one with a GRU RNN decoder as in the original work, and the other with a Transformer decoder. 
Table~\ref{tab:transformer-dialog} shows the results. Again, we see that the Transformer model generalizes well to the conversation generation setting, and consistently outperforms the GRU RNN counterpart. Regarding implementation efforts, our implementation based on Texar has around 100 lines of code, while the reference code~\citep{zhao2017learning} based on TensorFlow involves over 600 lines for constructing the same part.

\begin{table}[t]
%\begin{minipage}{.45\textwidth}
\small
\centering
\begin{tabular}{l l l}
\cmidrule[\heavyrulewidth]{1-3}
{\bf Models} & {\bf Test PPL} & {\bf Lines of Model Code} \\
\cmidrule{1-3}
LSTM RNN with MLE~\citep{zaremba2014recurrent} & 74.23 & 42 \\
LSTM RNN with seqGAN~\citep{yu2017seqgan} & 74.12 & 115 \\
Memory Network LM~\citep{sukhbaatar2015end} & 94.82 & 39 \\
\cmidrule[\heavyrulewidth]{1-3}
\end{tabular}
\vspace{-6pt}
\caption{Comparison of the three models on the task of language modeling, using the PTB dataset~\citep{zaremba2014recurrent}. The lower the test perplexity (PPL), the better. We also report the lines of code for implementing the respective model (including model construction, and loss/optimization specification, excluding data reading). In comparison, the official implementation of SeqGAN on TensorFlow involves 480 lines of model code~\citep{yu2017seqgan}. Texar implementations have limited the maximum length of each line to 80 chars.}
\label{tab:lm}
%\vspace{-25pt}
%\end{minipage}
\end{table}
%\hfill

\subsection{One Task with Many Techniques: Language Modeling}\label{sec:exp-one-task}
We next showcase how Texar can support investigation of diverse techniques on a single task. This can be valuable for research community to standardize experimental configurations and foster fair,  reproducible comparisons. As a case study, we choose the standard language modeling task~\citep{zaremba2014recurrent}. Note that this is different from the VAE language modeling task above, due to different data partition strategies conventionally adopted in respective research lines. 

We compare three models as shown in Table~\ref{tab:lm}. The LSTM RNN trained with maximum likelihood estimation (MLE)~\citep{zaremba2014recurrent} is the most widely used model for language modeling, due to its simplicity and prominent performance. We use the exact same architecture as generator and setup a (seq)GAN~\citep{yu2017seqgan} framework to train the language model with adversarial learning. (The generator is pre-trained with MLE.) From Table~\ref{tab:lm} we see that adversarial learning (almost) does not improve in terms of perplexity (which can be partly because of the high variance of the policy gradient in seqGAN learning). 
We further evaluate a memory network-based language model~\citep{sukhbaatar2015end} which has the same number of parameters with the LSTM RNN model. The test set perplexity is significantly higher than the LSTM RNNs in our experiments, which is not unexpected because LSTM RNN models are well studied for language modeling and a number of optimal modeling and optimization choices are already known.

Besides the benchmark performance of the various models, Table~\ref{tab:lm} also reports the amount of code for implementing each of the models. Texar makes the development very efficient.

%\begin{minipage}{.52\textwidth}
\begin{table}[t]
\small
\centering
\begin{tabular}{l l l l}
\cmidrule[\heavyrulewidth]{1-4}
{\bf Models} & {\bf Accuracy} & {\bf BLEU} & {\bf Lines of Model Code} \\
\cmidrule{1-4}
\citet{shen2017style} &  79.5  & 12.4 & 485 \\ \cmidrule{1-4}
\citet{shen2017style} on Texar & 82.5 & 13.0 & 136 \\
\citet{hu2017controllable} on Texar & {\bf 88.6} & {\bf 38.0} & 105 \\
\cmidrule[\heavyrulewidth]{1-4}
\end{tabular}
\vspace{-6pt}
\caption{Text style transfer on the Yelp data~\citep{shen2017style}. The first row is the original open-source implementation by the authors of \citep{shen2017style}. The subsequent two rows are Texar implementations of the two work. Texar implementations have limited the maximum length of each line to 80 chars.}
\label{tab:style}
%\vspace{-10pt}
%\end{minipage}
\vspace{-10pt}
\end{table}

\subsection{Composite Model Architectures: Text Style Transfer}\label{sec:exp-style}
To further demonstrate the versatility of Texar for composing complex model architectures, we next choose the newly-emerging task of text style transfer~\citep{hu2017controllable,shen2017style}. The task aims to manipulate the text of an input sentence to change from one attribute to another (e.g., from positive sentiment to negative), given only non-parallel training data of each style. The criterion is that the output sentence accurately entails the target style, while preserving the original content and other properties well. 

We use Texar to implement the models from both \citep{hu2017controllable} and \citep{shen2017style}, whose model architectures fall in the category (f) and (g) in Figure~\ref{fig:archs}, respectively. Experimental settings mostly follow those in \citep{shen2017style}. 
Following previous setting, we use a pre-trained sentiment classifier to evaluate the transferred style accuracy. For evaluating how well the generated sentence preserves the original content, we measure the BLEU score between the generated sentence and the original one (the higher the better)~\citep{yang2018unsupervised}. 
%Note that we do not mean to perform exhaustive evaluations of the methods, but instead aim to demonstrate the flexibility of the toolkit for implementing different composite model architectures beyond conventional encoder-decoder.
Table~\ref{tab:style} shows the results. Our re-implementation of \citep{shen2017style} replicates and slightly surpasses the original results, while the implementation of \citep{hu2017controllable} provides better performance in terms of the two metrics. Implementations on Texar use fewer lines of code for the composite model construction.

\section{Related Work}\label{sec:related}
Text generation is a broad research area with rapid advancement.
%(e.g., Figure~\ref{fig:archs} summarizes some popular and emerging models used in the diverse contexts of the field).
There exist several toolkits that focus on one or a few specific tasks. For example, for neural machine translation and alike, there are Google Seq2seq~\citep{britz2017massive} and Tensor2Tensor~\citep{vaswani2018tensor2tensor} on TensorFlow, OpenNMT~\citep{klein2017opennmt} on (Py)Torch, XNMT~\citep{neubig2018xnmt} on DyNet, Nematus~\citep{sennrich2017nematus} on Theano, MarianNMT~\citep{junczys2018marian} on C++, and others. For dialogue, ParlAI~\citep{miller2017parlai} is a software platform specialized for research in the field. Differing from these highly task-focusing toolkits, Texar aims to cover as many text generation tasks as possible. The goal of versatility poses unique challenges to the design, requiring high modularity and extensibility.

On the other end of spectrum, there are libraries and tools for more general natural language processing (NLP) or deep learning applications. For example, AllenNLP~\citep{allennlp}, QickNLP~\citep{quicknlp}, GluonNLP~\citep{gluonnlp} and others are designed for the broad NLP tasks in general, while Keras~\citep{chollet2017keras} is for high conceptual-level programming without specific task focuses. 
In comparison, though extensible to broader techniques and applications, Texar has a proper focus on the text generation sub-area, and provide a comprehensive set of 
modules and functionalities that are well-tailored and readily-usable for relevant tasks. For example, Texar provides rich \texttt{text docoder} modules with optimized interfaces to support over ten decoding methods (see section~\ref{sec:plugin} for an example), all of which can be invoked and interact with other modules conveniently.

It is also notable that some platforms have been developed for specific types of algorithms, 
such as OpenAI Gym~\citep{brockman2016openai}, DeepMind Control Suite~\citep{tassa2018deepmind}, and ELF~\citep{tian2017elf} for reinforcement learning in game environments. Texar has drawn inspirations from these toolkits when designing relevant specific algorithm supports.

%There are also libraries for general NLP applications~\citep{allennlp,quicknlp,gluonnlp} or for high conceptual-level programming without specific task focuses~\citep{chollet2017keras}. With the focus on text generation, we provide a more comprehensive set of well-tailored and readily-usable modules and functionalities to relevant tasks.
%Some platforms exist for specific types of algorithms, such as OpenAI Gym~\citep{brockman2016openai}, DeepMind Control Suite~\citep{tassa2018deepmind}, and ELF~\citep{tian2017elf} for reinforcement learning in game environments. Texar has drawn inspirations from these toolkits when designing relevant specific algorithm supports.

\section{Conclusion}
This paper has introduced Texar, an open-source, general-purpose toolkit on both TensorFlow and PyTorch, that supports the broad set of machine learning, especially text generation, applications and algorithms. The toolkit is modularized to enable easy replacement of components, and extensible to allow seamless integration of any external or customized modules. We are excited to further enrich the toolkit to support a broader set of natural language processing and machine learning applications.
%We invite researchers and practitioners to join and further enrich the toolkit, and in the end help push forward the text generation research and applications 

%\clearpage
%\small
\bibliography{refs}
\bibliographystyle{abbrvnat}

\end{document}